\documentclass[runningheads]{llncs}
\usepackage{times}
\usepackage{graphicx}
\usepackage{latexsym}
\usepackage{amsmath}
\usepackage{amssymb}
\usepackage{listings}
\usepackage{booktabs}
\usepackage{bookmark}
\usepackage{enumitem}
\usepackage{tikz}
\usepackage{tcolorbox}
\usepackage{lipsum}
\usepackage[utf8]{inputenc}
\usepackage{csquotes}

\xdefinecolor{pastel1}{RGB}{255, 252, 247}
\xdefinecolor{pastel2}{RGB}{255, 244, 228}
\xdefinecolor{pastel3}{RGB}{240, 220, 190}
\xdefinecolor{pastel4}{RGB}{205, 177, 136}
\xdefinecolor{pastel5}{RGB}{172, 139, 89}
\xdefinecolor{OuterFill}{RGB}{243,243,243}
\xdefinecolor{OuterBorder}{RGB}{130,130,130}

\colorlet{ground}[RGB]{pastel1!80!orange}
\colorlet{descriptive}[RGB]{pastel1!80!blue}

\usetikzlibrary{calc,shapes,positioning,arrows}
\tikzset{
  query/.style={draw=blue!10,thick,fill=blue!2,inner sep=.15cm},
  answer/.style={rectangle,draw=black!10,fill=gray!4},
  icon/.style={circle,thick,fill=blue!20,draw=blue!30,inner sep=.05cm,font={\bfseries \Large}},
  relation/.style={-latex,ultra thick},
  implies/.style={double,double equal sign distance,-implies,double distance=1mm},
  outer/.style={draw=OuterBorder,fill=OuterFill,thick,inner sep=5pt},
  physical/.style={fill=ground,draw=ground!80!black},
  social/.style={fill=descriptive,draw=descriptive!80!black},
}

\pgfdeclarelayer{background}
\pgfdeclarelayer{foreground}
\pgfsetlayers{background,main,foreground}   

\lstdefinelanguage[OWL]{XML} {
  basicstyle = \ttfamily,
  morekeywords={Individual,ObjectProperty,Types,Facts,Class,SubClassOf,Domain,Range,SubPropertyOf,EquivalentTo,exactly,some,only,and},
  morestring = [b]'
}
\lstdefinestyle{OWL} {language=[OWL]XML,
  lineskip=0.2ex,
  fontadjust=true,
  basicstyle={\scriptsize \nopagebreak[4]},
  stringstyle=\color{black!40!blue}\bf,
  showstringspaces=false,
}

\newcommand{\concept}[1]{\texttt{#1}}
\newcommand{\relation}[1]{\emph{#1}}

\hypersetup{
    pdftitle={Foundations of SOMA},
    pdfauthor={D. Be{\ss}ler et al.},
    pdfsubject={Activity Modeling},
    pdfkeywords={Knowledge Representation, Autonomous Robotics}
}

\newcommand{\givenODPNAME}{}
\newcommand{\givenODPINTENT}{}
\newcommand{\givenODPDEFINEDIN}{}
\newcommand{\givenODPDESCRIPTION}{}
\newcommand{\givenODPGRAPHIC}{}
\newcommand{\givenODPDOMAIN}{}
\newcommand{\givenODPQUESTION}{}

\newcommand{\OPDinit}{
  \renewcommand{\givenODPINTENT}{REQUIRED!}
  \renewcommand{\givenODPDEFINEDIN}{REQUIRED!}
  \renewcommand{\givenODPDESCRIPTION}{REQUIRED!}
  \renewcommand{\givenODPGRAPHIC}{REQUIRED!}
  \renewcommand{\givenODPQUESTION}{}
  \renewcommand{\givenODPDOMAIN}{}
}

\tikzset{owlclass/.style={draw=pastel4!60!black,fill=pastel1,rounded corners}}
\tikzset{ontobranch/.style={draw=none,rounded corners,minimum width=2.3cm,minimum height=0.6cm,text width=2.0cm, align=center}}
\tikzset{groundbranch/.style={ontobranch,physical}}
\tikzset{socialbranch/.style={ontobranch,social}}

\newenvironment{owlclass}[2][,] {
  \begin{minipage}{5.0cm}
  \begin{center}
  \texttt{\bf#2} \\[-0.2cm]
  \par\noindent\textcolor{pastel4!80!black}{\rule{\textwidth}{0.4pt}}
  \vspace{-0.6cm}
  \begin{itemize}[#1]
  \raggedright
  } {
  \end{itemize}
  \end{center}
  \end{minipage}
}

\newcommand{\somamodule}[3]{
    \begin{minipage}{4.8cm}
        \begin{center}
            \texttt{\bf #1} \\[-0.2cm]
            \par\noindent\textcolor{pastel4!80!black}{\rule{\textwidth}{0.4pt}}
            
            \vspace{4pt}\noindent\begin{tikzpicture}
            \node[groundbranch] (LEFT) {#2};
            \node[socialbranch,right=0.1cm of LEFT] {#3};
            \end{tikzpicture}\vspace{2pt}
        \end{center}
    \end{minipage}
}

\OPDinit

\begin{document}
\title{Foundations of the Socio-physical Model of Activities (SOMA) for Autonomous Robotic Agents\thanks{This work was funded by the German Research Foundation (DFG) as part of Collaborative Research Center (SFB) 1320 EASE -- Everyday Activity Science and Engineering, University of Bremen (\url{http://www.ease-crc.org/}), subprojects H2, P1 and R1.}}
\titlerunning{Foundations of SOMA}
\author{Daniel Be{\ss}ler\inst{1} \and
Robert Porzel\inst{2} \and
Mihai Pomarlan\inst{3} \and
Abhijit Vyas\inst{1} \and
Sebastian H{\"o}ffner\inst{2} \and  
Michael Beetz\inst{1} \and
Rainer Malaka\inst{2} \and
John Bateman\inst{3}}
\authorrunning{D. Be{\ss}ler et al.}
%
\institute{Department of Artificial Intelligence,
        University of Bremen, 28359 Bremen, Germany \and
Digital Media Lab,
        University of Bremen, 28359 Bremen, Germany \and
Department of Linguistics,
        University of Bremen, 28359 Bremen, Germany
}
\maketitle

\begin{abstract}
In this paper, we present foundations of the Socio-physical Model of Activities (SOMA). SOMA represents both the physical as well as the social context of everyday activities. Such tasks seem to be trivial for humans, however, they pose severe problems for artificial agents. For starters, a natural language command requesting something will leave many pieces of information necessary for performing the task unspecified. Humans can solve such problems fast as we reduce the search space by recourse to prior knowledge such as a connected collection of plans that describe how certain goals can be achieved at various levels of abstraction. Rather than enumerating fine-grained physical contexts SOMA sets out to include socially constructed knowledge about the functions of actions to achieve a variety of goals or the roles objects can play in a given situation. As the human cognition system is capable of generalizing experiences into abstract knowledge pieces applicable to novel situations, we argue that both physical and social context need be modeled to tackle these challenges in a general manner. This is represented by the link between the physical and social context in SOMA where relationships are established between occurrences and generalizations of them, which has been demonstrated in several use cases that validate SOMA.
\end{abstract}

\section{Introduction}

In spite of undoubtedly being ubiquitous, the domain of \emph{everyday activities} poses considerable challenges.
Many people perform activities such as cooking or cleaning almost every day.
This includes to
select and manipulate ingredients, use tools and devices, arrange the prepared dishes for serving and clean up afterwards -- and do it all quick and robust without recourse to an advanced computational theory.
Further, the amount of information provided in a description of a task -- such as a natural language command requesting its completion -- is much less than the amount of information needed to perform the task. This 
raises the question how humans are able to decide so quickly what to do next, despite ambiguity and underspecification.

Lenat and Feigenbaum observe that \enquote{more knowledge implies less search}~\cite{lenat91}. Knowledge of many possible plans, as well as knowledge of the world in general, seems to be the secret of human performance. There is no algorithmic reason why tomatoes and oregano go well together, or why a raw egg must be handled with care. The cook simply has to know these things. Such knowledge of the world is taught, observed, and then ingrained by practice. As Anderson observes, \enquote{an agent has a great deal of knowledge [of everyday activities], which comes as a result of the activity being common}~\cite{anderson95}. As human beings, we acquire such knowledge naturally over the course of our lives.
Furthermore, to provide a better understanding of our subsequent discussion,
we will briefly introduce some of the essential concepts and data structures that drive our research efforts.
A lot of what we learn, we learn by doing, or by watching others. This suggests that a robot must have mechanisms to organize and interpret observations, either of its own behavior or of other agents, into structures that are then amenable for other computational tasks. To this end, the EASE-CRC put forth the concept of \emph{narratively-enabled episodic memories} (NEEMs) as a representation of the physical context. NEEMs are comprehensive logs of raw sensor data, actuator control histories and perception events, all semantically annotated with information about what the robot is doing and why using the terminology provided by SOMA. 

The computational tasks that must be solved when acting in the physical world are often very complex and beyond what is thought to be tractable. This, however, is only the case when these problems are regarded in their full generality and not for restricted versions of these problems. However, the knowledge representing such pragmatic solutions goes beyond modeling physical events and requires models of the social context by means of which the physical events can be realized and interpreted.
To represent this, the EASE-CRC has put forth the concept of \emph{pragmatic everyday activity manifolds} (PEAMs), which are descriptions of a problem subspace that are more amenable to solving. Depending on the larger problem kind,  PEAMs can, for example, be expressed as plans with some placeholders for subtasks, using vague parameterizations for controllers and heuristic strategies to classify objects by their  dispositions.

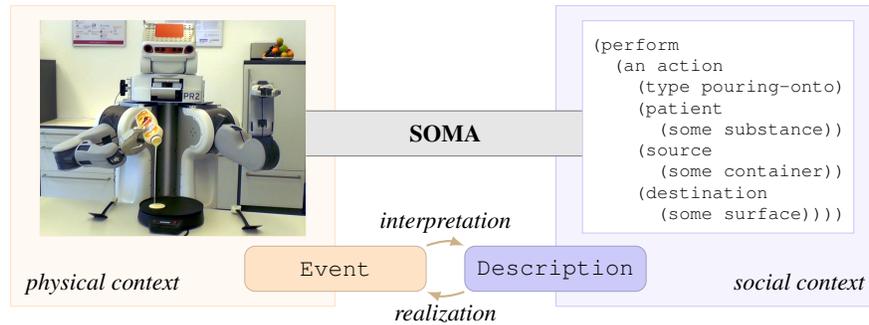
\begin{figure}[t]
    \lstset{basicstyle=\scriptsize\ttfamily,breaklines=true}
    \centering
    \begin{tikzpicture}[
            myedge/.style={thick,pastel4,-latex},
            mylabel/.style={black,font=\small\it},
            myimage/.style={},
            mynode/.style={draw=none,minimum width=2.4cm,minimum height=0.6cm,text width=2.2cm, align=center},
            myconcept/.style={mynode,rounded corners},
            mysection/.style={draw=red,minimum width=4.1cm,minimum height=4.0cm,text width=4.1cm, align=center},
        ]
        \node[myconcept,physical] (EV) {\concept{Event}};
        \node[myconcept,social,right=0.5cm of EV] (DESCR) {\concept{Description}};
        \draw (EV.north east) edge[myedge,bend left=20]
                              node[mylabel,above] {interpretation}
                              (DESCR.north west);
        \draw (DESCR.south west) edge[myedge,bend left=20]
                                 node[mylabel,below] {realization}
                                 (EV.south east);
        \begin{pgfonlayer}{background}
        \node[mysection,left=0.0cm of EV.center,yshift=1.5cm,
              fill=ground!20!white,draw=ground] (LEFT) {};
        \node[mysection,right=0.0cm of DESCR.center,yshift=1.5cm,
              fill=descriptive!20!white,draw=descriptive] (RIGHT) {};
        \node[mylabel,anchor=west,yshift=-0.2cm,xshift=0.1cm] at (EV.west -| LEFT.west) (PC) {physical context};
        \node[mylabel,anchor=north east,xshift=-0.1cm] at (PC.north east -| RIGHT.east) {social context};
        \end{pgfonlayer}
        \node[below=0.2cm of RIGHT.north,anchor=north,
              fill=white,draw=descriptive] {
            \begin{lstlisting}
(perform
  (an action
    (type pouring-onto)
    (patient
      (some substance))
    (source
      (some container))
    (destination
      (some surface))))
            \end{lstlisting}
        };
        \node[below=0.2cm of LEFT.north,anchor=north,inner sep=0,draw=ground] {
            \includegraphics[trim=30px 60px 20px 70px,clip,width=100px]{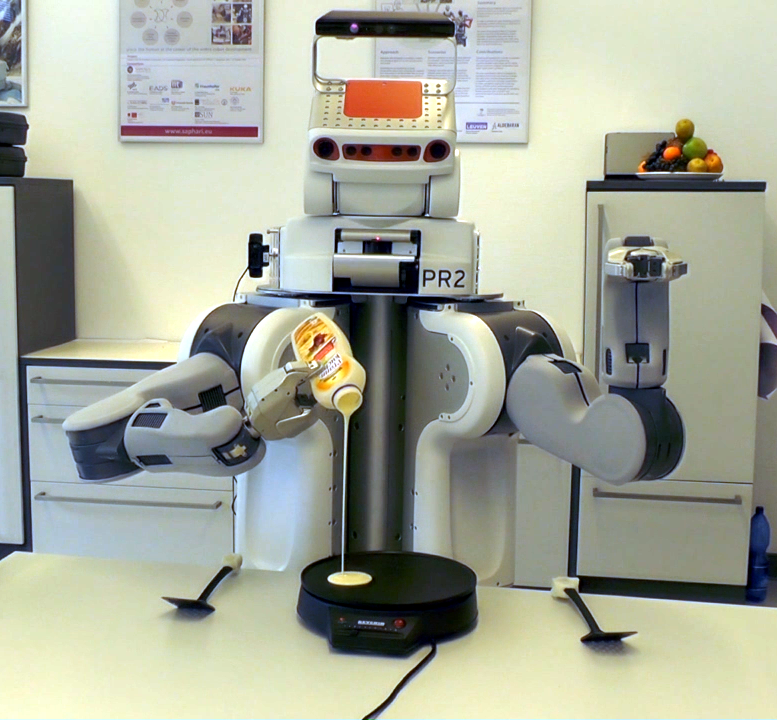}
        };
        \begin{pgfonlayer}{background} 
        \node[mynode,fill=gray!20!white,draw=gray,yshift=1.8cm,text width=6.2cm] (SOMA) at ($(EV)!0.5!(DESCR)$) {\textbf{SOMA}};
        \end{pgfonlayer}
    \end{tikzpicture}
    \caption{SOMA represents physical and social context, and supports robotic agents in interpreting observed events, and realizing abstract descriptions.}
    \label{fig:intro}
\end{figure}

The overall goal of our research is to enable robotic agents to perform everyday activities with similar robustness and flexibility as human agents do. Given this aim of the EASE-CRC, we must, in some sense, get the robot to know what humans know about the world, at least as it pertains to everyday activities. This presents several challenges, beyond the scope of what needs to be known to represent such intricate and extensive domain.
There lies the question of how to represent and structure this knowledge in order to realize a similar robustness, flexibility and efficiency in performance.
In addition, there are challenges concerning
the acquisition and learnability of the corresponding structures. In this paper, we will focus on how this knowledge is represented. 
We will describe an upper level ontology together with several ontology modules aimed to address this general ontology design challenge.
The ontology is openly available~\footnote{\url{https://github.com/knowrob/ease_ontology/tree/ekaw202}}.
Individual applications of our model have been published in dedicated venues and will be referenced and summarized to validate the contribution of this work.
As depicted in Figure~\ref{fig:intro}, our focus lies on representing both the physical context of realized everyday activities (NEEMs), as well as interpretations thereof as the social context (PEAMs).

\section{Related Work}

Knowledge representation and reasoning in autonomous robot control is a fairly extensive field of research with developments in both service and industrial robotics.
Olivares et al. provide a comprehensive comparison of different approaches~\cite{Olivares19}.

One example in the industrial robotics domain is the ROSETTA project~\cite{Patel2012,Malec2013,Stenmark2015}.
Its initial scope was reconfiguration and adaptation of robot-based manufacturing cells, however, the authors have, since then, further developed their activity modeling for coping with a wider range of industrial tasks.
Other authors have focused on modeling
industrial task structure, part geometry features, or task teaching from examples~\cite{Balakirsky2013,Balakirsky2015,PolydorosGRNK16,Kootbally2015,PerzyloSPKRK16}.
Compared to the everyday activity domain, industrial tasks considered in above works are more structured, and less demanding in terms of flexibility.

An approach to activity modeling in the service robotics domain is presented by Tenorth and Beetz~\cite{tenorth2015representations}.
The scope of their work is similar to ours as the authors also consider how activity knowledge can be used to fill knowledge gaps in abstract instructions given to a robotic agent performing everyday activities.
However, the scope of the work presented here is wider, as we also consider how activity knowledge can be used for the interpretation of observations.
Our activity modeling is further more detailed in terms of activity structure as we also consider the processes and states that occur during an activity.
Another difference is that, in their modeling, there is no distinction between physical and social context, and therefore less expressivity compared to our model.

A more general approach to activity modeling for robotic agents is presented by the IEEE-RAS working group ORA~\cite{schlenoff2012ieee}.
The group has the goal of defining a standard ontology for various sub-domains of robotics, including a model for object manipulation tasks. 
It has defined a core ORA ontology~\cite{PRESTES20131193}, as well as additional modules for industrial tasks such as kitting~\cite{Fiorini2015}.
In terms of methodology, we differ in foundational assumptions we assert, which has important consequences on the structure of our ontology, modeling workflow, and inferential power.
In the case of ORA, the SUMO upper-level ontology is used as foundational layer.
Compared to SUMO, we use a richer axiomatization of entities on the foundational layer,
and put particular emphasis on the distinction between physical and social activity context.

\section{Overview}
\label{sec:overview}

In this section, we will provide a broad overview about SOMA.
First, we will discuss the scope of our work which is rooted in autonomous robotics in Section~\ref{subsec:scope}.
We will then elaborate on ontological commitments in SOMA, and in particular about commitments that help coping with both physical and social context in Section~\ref{subsec:commitments}.
Finally, we will provide an overview about how SOMA is organized into different ontological modules in Section~\ref{subsec:modules}.

\subsection{Scope} 
\label{subsec:scope}

The broad scope of our work is everyday object manipulation tasks in autonomous robot control, and in particular the motion and force characteristics of objects that interact with each other.
The research question driving us is whether a single general control program can be written that can generate adequate behavior in many different contexts: for different tasks, objects, and environments.

One of the challenges is that, using such a general plan, the agent needs to fill the knowledge gaps between abstract instructions included in the plan and the realization of context specific behavior. That is, for example, the many ways of how humans perform a pouring task depending on the source from which is poured, the destination, and the substance that is to be poured.
Another challenge is that object manipulation tasks may fail if the agent does not perform the motions competently and well. This is caused by the agent choosing inappropriate parametrization of its control-level functions.

The employment of a general plan thus requires an abstract task and object model, and a mechanism to apply this abstract knowledge in situational context.
To achieve this, an agent needs to be equipped with the necessary common-sense and intuitive physics knowledge, which is what SOMA attempts.

\subsection{Foundational Commitments}
\label{subsec:commitments}

We decided to base our model on the DOLCE+DnS Ultralite (DUL) foundational framework~\cite{DOLCE2003}.
This decision is greatly motivated by their underlying ontological commitments.
Firstly, DUL is not a revisionary model, but seeks to express stands that shape human cognition. Furthermore it assumes a reductionist approach -- rather than capturing, for example, the flexibility of our usage of objects via multiple inheritance in a multiplicative manner, we commit to a reduced {\it ground} classification and use a {\it descriptive} approach for handling this flexibility. For this a primary branch of the ontology represents the ground {\bf physical model}, e.g. objects and actions, while a secondary branch represents the {\bf social model}, e.g. roles and tasks. All entities in the social branch would not exist without humans, i.e. they constitute social objects that represent concepts about or descriptions of ground elements. 

Every axiomatization in the physical branch can, therefore, be regarded as expressing some physical context whereas axiomatizations in the descriptive social branch are used to express social contexts. A set of dedicated relations is provided that connect both branches. For example, as detailed in Section~\ref{subsec:roles}, the relation \emph{classifies} connects ground objects, e.g. a hammer, with the roles they can play, i.e. potential classifications. Thus, we can state that a hammer can in some context be conceptualized as a murder weapon, a paper weight or a door stopper. Nevertheless, neither its ground ontological classification as a tool will change nor will hammers be subsumed as kinds of door stoppers, paper weights or weapons via multiple inheritance. Following a quick overview of the central modules of SOMA where these commitments apply, we will provide detailed examples of where and how our commitments apply in Sections~\ref{sec:perdurants} and~\ref{sec:endurants}.

\subsection{Module overview}
\label{subsec:modules}

\begin{figure}[t]
    \centering
    \begin{tikzpicture}[myedge/.style={thick,-,pastel4}]
        \node[owlclass] (SOMA)
        { \somamodule{SOMA}{physical context}{social \ \ \ \ context} };
        \node[owlclass,below right=1.0cm of SOMA,xshift=-2.5cm,yshift=0.4cm] (OBJ) { \somamodule{OBJ}{Object, Disposition}{Design, Affordance} };
        \node[owlclass,below left=1.0cm of SOMA,xshift=2.5cm,yshift=0.4cm] (EXEC) { \somamodule{EXEC}{Grounding}{Binding} };
        \node[owlclass,above=2.5cm of SOMA,yshift=-0.4cm] (ACT)
        { \somamodule{ACT}{Action}{Plan} };
        \node[owlclass,above left=1.0cm of SOMA,xshift=2.5cm,yshift=-0.4cm] (PROC) { \somamodule{PROC}{Process}{Progression} };
        \node[owlclass,above right=1.0cm of SOMA,xshift=-2.5cm,yshift=-0.4cm] (STATE)
        { \somamodule{STATE}{State}{Configuration} };
        \draw (SOMA) edge[myedge] (OBJ);
        \draw (SOMA) edge[myedge] (EXEC);
        \draw (SOMA) edge[myedge] (ACT);
        \draw (SOMA) edge[myedge] (PROC);
        \draw (SOMA) edge[myedge] (STATE);
    \end{tikzpicture}
    \caption{The modular organization of SOMA. Each module defines concepts and relationships used to represent physical (orange) and social (purple) activity context.}
    \label{fig:soma_model_overview}
\end{figure}
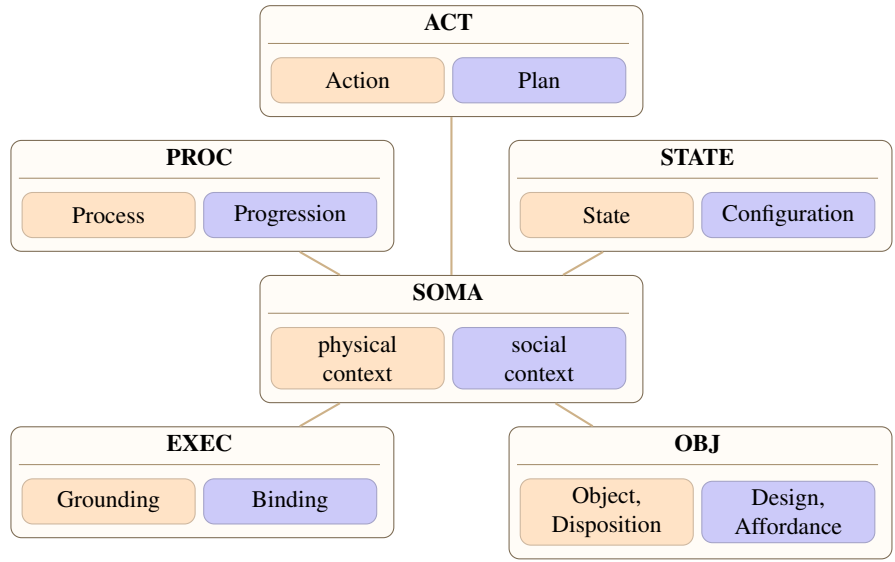

SOMA is organized in several modules that conceptualize different aspects of physical and social activity context (Figure~\ref{fig:soma_model_overview}). The different modules correspond to different event types (\emph{ACT}, \emph{PROC}, \emph{STATE}), objects that participate in the activity (\emph{OBJ}), and execution context (\emph{EXEC}). The scope of each of these modules is outlined below.

\paragraph{OBJ.}
The scope of the \emph{OBJ} module is the contextualization of physical objects, and their qualities. The module includes two taxonomies used to classify objects: an object taxonomy in the grounded branch, and a role taxonomy in the descriptive branch. It further includes a taxonomy of dispositions to represent the potential of using an object in some way, and a design taxonomy used to categorize objects based on function, structure, and aesthetics.

\paragraph{ACT.}
The scope of the \emph{ACT} module is the contextualization of actions. Actions are events performed by an agent (physical context), and structured by a plan that is executed by the agent (social context). Primarily, this module defines a taxonomy of actions in the descriptive branch. It further includes concepts and relations used to represent the structure of an action, and the situational context of its execution.

\paragraph{PROC.}
The purpose of the \emph{PROC} module is the contextualization of processes. Processes are interpreted as events that are considered in their evolution. This is, e.g., the case for motions. A process taxonomy is defined in this module, as well as concepts that describe their evolution, and situational context.

\paragraph{STATE.}
States are the third type of events considered in our modeling.
For example, objects being in contact with each other is interpreted as a state.
The difference between a state and a process is that, when considering time slices  of the event, for states, these time slices always have the same type as the state (states are homeomeric), but for processes this is not the case.
The \emph{STATE} module defines a state taxonomy, a general concept used to conceptualize states, and concepts used to represent situational context.

\paragraph{EXEC.}
The \emph{EXEC} module covers representations of execution contexts. It defines concepts related to parameterizing and connecting tasks. It provides ways to describe how outcomes of some tasks relate to inputs of others, and therefore characterize how tasks depend on, or enable one another.

\section{Object Representation in SOMA}
\label{sec:endurants}

One of the reasons that everyday activity is such a hard problem is the immense amount of variations an unrestricted environment may have, and the resulting potentials of interaction for an agent. Each type of object needs to be handled differently depending on its properties.
However, object manipulation tasks are often defined independent of the type of object that is manipulated. It is thus crucial for an agent that copes with these variations to have an abstract object model, and a mechanism for applying abstract object knowledge to novel situations.

The main link between physical objects and activities in SOMA is that objects participate in events in the physical branch, and that the social branch represents the interpretation of their participation. This is elaborated in Section~\ref{subsec:roles}.
Next, in Section~\ref{subsec:design}, we will discuss how objects are organized along their design, and how design knowledge is used to cope with general tasks.
However, the agent might further need to find suitable candidate objects to perform a task by reasoning about which objects have the potential to be used in a certain way.
We employ an object disposition model for that purpose which is discussed in Section~\ref{subsec:disposition}.

\subsection{Object Types}
\label{subsec:roles}

For the classification of objects we employ the \texttt{Role} pattern provided by the foundational layer. Roles are \texttt{Concepts} and, as such, reside in the \texttt{SocialObject} branch of DUL. For human agents the ascription of roles to entities comes very natural. {\it He is a student} does not imply an \emph{isa} or \emph{instanceof} relation between some male individual and a student class. It is rather meant that at this point of his life the individual plays the role of a student, which, however, can and will change over time. Therefore, a \emph{classifies} relation is used to express that someone can be classified as a student or something can be classified as a container.

This role pattern is of paramount importance, especially in the modeling of affordance discussed below in Section~\ref{subsec:disposition}. In the model presented herein, we import the roles that have been established in the field of frame semantics \cite{Baker98}. The selectional restrictions imposed by the \emph{classifies} relation are used in a number of reasoning processes ranging from natural language understanding to tool selection. As certain roles can only classify physical agents or specific types of designed artifacts these axiomatizations provide substantial information about context dependent {\it meaning} of objects.

\subsection{Object Designs}
\label{subsec:design}

The organization of objects along a taxonomy is difficult as objects can be categorized in many ways. A notion of design is useful to capture object categories corresponding to structural, functional, or aesthetic patterns.
Designs are in particular useful to conceptualize \emph{refunctionalized} entities, and to support an agent to \emph{hypothesize} unknown functions served by an entity. For example, a wooden pallet can be reused for the construction of furniture such as sofa, bed etc. The categorization of objects along their design can be employed in order to allow the use of more general plans, where, instead of object types, the plan refers to structure, aesthetics, or function.

Within the scope of SOMA, the \texttt{Design} concept belongs to the social branch. SOMA considers \emph{structural}, \emph{functional} and \emph{aesthetic} aspects of design. A design describes classes of objects that host a common design-relevant quality. These qualities are dispositional, geometrical, and aesthetic aspects of the object. This corresponds to our design categorization into functional, structural, and aesthetic design. Each \texttt{Design} concept defines restrictions on the corresponding quality type that needs to be fulfilled by any object described by the design. These restrictions also represent sufficient conditions under which an object is thought to be described by the design which allows the classification of entities given their design pattern can be detected.

In the scope of this work, we only consider functional aspects of objects. These are represented using a model of dispositional qualities which is discussed next.

\subsection{Object Dispositions}
\label{subsec:disposition}

Objects are important to an agent because they allow it to perform, or prevent it from performing, actions to achieve its goals. The notion of ``affordance'' was put forth by Gibson to cover this concept~\cite{gibson1979ecological}: 

\begin{quote}``The affordances of the environment are what it offers the animal, what it provides or furnishes, either for good or ill.''\cite{gibson1979ecological}\end{quote}


However, though evidently useful as a way to organize actionable knowledge about the world~\cite{Wan18}, affordances proved very difficult to model ontologically. Several approaches have been proposed, such as regarding affordances as qualities~\cite{Ortmann10} or as events~\cite{moralez2016affordance}.
Nonetheless, we think these approaches are not satisfactory. Affordances are relational, characterizing a potential interaction of several objects, and therefore should not be treated as either a quality belonging to an object, nor as an event. We do recognize that some qualitative aspects of objects contribute to affordances, however, which is why we constructed our model around the interplay of Turvey's notion of \emph{disposition}~\cite{turvey1992ecological} and Gibson's notion of \emph{affordance}.

In SOMA, the \concept{Disposition} concept is defined as an object quality that allows an object to
participate in events that \emph{realize} an affordance.
The \concept{Affordance} concept itself, however, is defined as the relational context holding between several objects that play different roles such as being the ``bearer'', ``trigger'', or ``background'' of an affordance.
Our modeling allows us, via a mixture of DL and other reasoning mechanisms such as simulation, to answer several interesting questions such as what affordances might an object provide in some combination with others, what objects might, or probably would not, be able to provide a given affordance, what combinations of objects would work towards providing an affordance etc.
For more details on our disposition and affordance model, we invite the reader to consult our previous work on this topic~\cite{bessler2020ecai}.

\section{Event Representation in SOMA}
\label{sec:perdurants}

The information gap between an instruction given to an embodied agent and the way it has to move its body to \emph{successfully} execute the instruction is often immense. Consider, for example, a recipe for cooking noodles that contains an instruction to \emph{boil water in a pot}. It is simple to decompose this instruction into several steps with individual sub-goals such as finding pot and tap, placing the pot underneath the tap, and filling the pot with water. However, the more difficult problem is how the agent has to move its body in each step such that the goal is achieved, and unwanted side-effects are avoided.
This is, for example, to avoid spillage of the water when the valve is opened too much, or when the agent moves too fast with the water-filled pot.

Little variations in motion behavior may have drastic consequences in tasks that require delicate interaction. It is thus essential for agents performing actions in the physical world to reason about \emph{how they should move} to achieve their goals in an appropriate, flexible and robust manner which is an unsolved problem for the general case.

SOMA attempts to support an agent facing this problem by equipping it with knowledge about relationships between abstract descriptions and their realization.
The support is twofold. First, the agent may employ more general plans where informational gaps are filled by reasoning over knowledge represented with SOMA. Second, the agent may employ SOMA for understanding and generalizing observations. This means that agents can interact safer in environments with incomplete information, and that they can learn general patterns from specific situations.
An illustrative example of the representation of a pouring plan in OWL Manchester Syntax is provided in Figure~\ref{fig:plan}.

\begin{figure}[t]
    \centering
\begin{tcolorbox}[notitle, colback=descriptive!20!white, colframe=descriptive, boxrule=0.2mm, arc=0.0mm, boxsep=0mm, top=1mm, bottom=0mm]
\begin{multicols}{2}
\begin{lstlisting}[style=OWL]
Individual: PouringPlan_0
    Types: Plan, Description
    Facts: defines Pouring_0,
           hasPhase Approaching_0,
           hasPhase Tilting_0
Individual: Pouring_0
    Types: Task, Concept
    Facts: usesRole Patient_0,
           usesRole Source_0,
           usesRole Destination_0,
           startedBy Approaching_0
\end{lstlisting}
\vfill
\begin{lstlisting}[style=OWL]
Individual: Approaching_0
    Types: Motion Type, Concept
    Facts: usesRole Destination_1,
           overlapsWith Tilting_0
Individual: Tilting_0
    Types: Motion Type, Concept
    Facts: usesRole Patient_1
Individual: Binding_1
    Types: RoleBinding, Description
    Facts: hasBinding Source_0,
           hasBinding Patient_1
\end{lstlisting}
\end{multicols}
\end{tcolorbox}
    \caption{An example of how plans are represented in SOMA.}
    \label{fig:plan}
\end{figure}

In the following, we will describe the representation of activities in SOMA in more detail. We will first introduce our hierarchical organization of tasks, processes and states in Section~\ref{subsec:eventtype}, and, second, how they are decomposed into phases with explicit goals and individual knowledge pre-conditions in Section~\ref{subsec:phases}. Finally, we will discuss our modeling of force characteristics
in Section~\ref{subsec:force}.

\subsection{Event Types}
\label{subsec:eventtype}

One of the most important demands on a cognitive system is to reason about actions; colloquially speaking, an agent constantly asks itself what to do, and how to do it. This however opens up another question, namely what exactly is the entity that the agent represents -- an actual event, or an interpretation of one.


As an example, consider this scenario: a robot moves toward a table carrying a plate. Midway, its gripper releases, dropping the plate, which shatters against the floor. Perhaps the robot had to transport the plate to the table, and it failed to do so; or perhaps it was required to drop the plate as part of some material test, and the table was just there for some other reason. Just by observation of the action,
without
other interpretive context which includes knowledge of what the robot was told to do, there is no reliable way to tell. The failed transport interpretation does seem more likely a priori, but only because we have more often seen people tell robots to transport plates rather than break them; we still make use of an expected interpretive context.

As a result, we do not define a taxonomy of action events in our ontology, but rather of tasks that are used to conceptualize actions.
For example, the \concept{Grasping} concept is defined as task in SOMA, and it is used for the \emph{classification} of events that are interpreted as an intentional grasping activity.
This classification pattern between events and their conceptualization is provided by the foundational layer of SOMA.
However, within the foundational layer, this pattern is only instantiated for actions and their conceptualization. In our modeling, motions of an agent and other processes, as well as state events are used to structure an activity. Thus, we also need to represent processes and states in the ground and the descriptive ontology.
The same pattern applies: the concepts \concept{Process} and \concept{State} are defined in the ground ontology, and their conceptualization in the descriptive ontology, and a relationship between both branches is established through the aforementioned classification pattern.



\subsection{Event Phases}
\label{subsec:phases}

Actions in SOMA are composed of distinct \emph{phases}. Each phase has its individual goal, and requires a different movement strategy to be executed successfully. The phases correspond to different stages of an object manipulation task, usually separated through \emph{contact events}.
Flanagan et al. have pointed out the importance of contact events in object manipulation tasks~\cite{flanagan06control}. The authors have shown that contact events cause a distinct pattern in sensory events, and that they can be used as \emph{sensorimotor control points} for aligning and comparing predictions with actual sensory events. 
Another justification is that humans have shown to direct their gaze to contact points when they perform object manipulation tasks, or when they observe another agent performing a task.

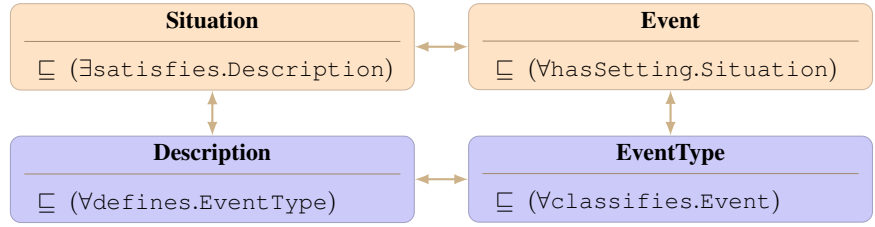
\begin{figure}[t]
    \centering
    \renewcommand{\labelitemi}{$\mathbf{\sqsubseteq}$}
    \begin{tikzpicture}[myedge/.style={thick,latex-latex,pastel4}]
        \node[owlclass,physical] (SIT) {
            \begin{owlclass}{Situation}
                \item $(\exists \texttt{satisfies}.\texttt{Description})$
            \end{owlclass}
        };
        \node[owlclass,social,below=0.6cm of SIT] (DESCR) {
            \begin{owlclass}{Description}
                \item $(\forall \texttt{defines}.\texttt{EventType})$
            \end{owlclass}
        };
        \node[owlclass,social,right=3.4cm of DESCR.north east,anchor=north] (CON) {
            \begin{owlclass}{EventType}
                \item $(\forall \texttt{classifies}.\texttt{Event})$
            \end{owlclass}
        };
        \node[owlclass,physical,right=3.4cm of SIT.north east,anchor=north] (EV) {
            \begin{owlclass}{Event}
                \item $(\forall \texttt{hasSetting}.\texttt{Situation})$
            \end{owlclass}
        };
        \draw (SIT) edge[myedge] (DESCR);
        \draw (DESCR) edge[myedge] (CON);
        \draw (CON) edge[myedge] (EV);
        \draw (EV) edge[myedge] (SIT);
    \end{tikzpicture}
    \caption{The "Interpretation" pattern used to link the physical branch of SOMA (top) with social context (bottom).}
    \label{fig:description}
\end{figure}

The structure of activities in SOMA is governed by a set of design patterns applied to different levels of granularity. At its core, SOMA activity modeling builds on top of the \emph{basic plan} ontology design pattern that represents plans and their execution.
The pattern further defines that an execution is a situation that \relation{satisfies} the description of the plan.
However, the pattern is defined too specific for the scope of this work, as we also want to provide descriptive context for states and processes.
Hence, we generalize this pattern such that it can be instantiated in different levels of granularity in the activity model. This generalized pattern is displayed in Figure~\ref{fig:description}. Its purpose is to link the grounded with the descriptive branch.

In SOMA, this pattern is instantiated for actions, states and processes:
\begin{itemize}
    \item An \concept{Action} is described in a \concept{Plan} which is a description having an explicit goal. A plan satisfies situations that include action sequences that match the structure of the plan, such situations are called \concept{Plan Executions};
    \item a \concept{State} is described in a \concept{Configuration} which includes constraints on regions of entities and relationships between them. A configuration satisfies situations in which all constraints of the configuration are satisfied; and
    \item a \concept{Process} is described in a \concept{Process Flow} which is a description of the progression of the process. A process flow satisfies situations that include a process that progresses in the described way.
\end{itemize}

SOMA represents the structure of an activity using a sequence pattern based on Allen's interval calculus~\cite{Allen83}. Allen's calculus defines thirteen relations between time intervals including \relation{before}, \relation{after}, \relation{overlaps}, and \relation{meets}. This is useful to, on the one hand, represent precedence of one phase strictly following the other, and, on the other hand, it allows to cope with concurrency in the sequence. We apply this algebra to event types that are defined within the descriptive context of a plan or process flow. However, reasoning about sequences is not well supported in OWL. Instead, we employ a specialized reasoner that internally builds a graph that encodes sequence relations, and which can be queried to infer whether some sequence relation exists between two entities or not.

Another aspect of activity structure can be captured by SOMA in, what we call, \emph{execution contexts}. These are representations of how different phases of an activity constrain each other depending on conditions encountered in the activity execution. In particular, we define the \texttt{Binding} concept as identity constrain representing that a parameter or role grounding is the same in different phases, however potentially being classified differently. 

Knowledge about the structure of activities can be employed by an agent in both directions: for planning an activity, and for interpreting observed events. Planning can be seen as a mapping from the descriptive to the grounded branch of SOMA, while interpretation maps the other way.
For embodied agents, planning goes beyond mere decomposition of an activity into steps, the agent may further need to decided what objects it should use, how it should move, with what speed, and how much force it should apply when getting into contact with some object.
SOMA may be employed by the agent to find potential sequences of steps and motions to execute a task, to support finding potential objects playing some role during the activity, and to constrain the values of parameters of a task.
Interpretation of events, on the other hand, is difficult because the intentions of other agents are not known. However, it is often possible to detect contact events, types of motions, and states reliably. These can  be used as tokens for an activity parser that uses SOMA as a grammar, this will be described in more detail in Section~\ref{sec:application}.

\paragraph{Knowledge Pre-Conditions.}
In order to execute a motion, an agent has to invoke one of its control routines with a set of arguments. Higher-level routines may have a notion of object, but at a lower-level all boils down to numbers such as with what effort the robot moves, how fast, etc. SOMA allows to define constraints for both cases: for the types of objects that can play a role during the action, and for the value of parameters. This is done by using restrictions on what types of objects or regions can be classified by some role or parameter. This information is used to reduce the search space for doing an appropriate object or parameter selection (Section~\ref{subsec:disposition}).

\paragraph{Goals.}
A goal is a description of a desired situation, and it is achieved only if the situational context, after the execution has been finished, satisfies this description. One would say then that the action was executed successfully.
SOMA is more specific about what it means to execute an action successfully as it decomposes it into processes and states where the goal of the task is that the progression of processes evolves, and that state changes occur as described.
Particularly important are the contact states in object manipulation tasks, as they represent control points for the agent when generating or observing behavior.

\subsection{Event Force Characteristics}
\label{subsec:force}

A contact state is an indicator for whether objects are touching each other or not. Patterns of such states are useful for distinguishing between categories of activities. However, different activities may cause the same pattern while their goal is different, or even the opposite of each other. This is, for example, the case for pulling and holding. Both cause the same pattern of an endeffector getting into contact with another object. But the force characteristics are different: the goal of a pulling task is to overcome the inertial force of the object to set it into motion, and the goal of a holding task is to neutralize any external force that would set the object into motion.
Another aspect is that an agent performing such a task needs to decide how much force to apply. In order to make this decision it is valuable to know what the intended force-related consequences are.

SOMA supports the representation of force characteristics using Talmy's notion of force dynamics~\cite{Talmy2000}. Talmy distinguishes between two entities that participate in force dynamical processes: the \concept{Agonist}, and the \concept{Antagonist}. An agonist is the subject of a force dynamical expression, while the antagonist is the opposing force in the expression. Each expression has an intrinsic force tendency either to set the agonist into motion, or to keep it resting. Whether the tendency can be realized or not depends on which of the two entities is the \emph{stronger} entity.

\section{Applications of SOMA}
\label{sec:application}

The applicability of SOMA in the domain of everyday activities in autonomous robot control has been demonstrated in several scientific publications. We will summarize some of them below in order to provide evidence that SOMA contributes to coping with the complexities of everyday activities~\footnote{please note that SOMA is referred to as "EASE ontology" or "NEEM narrative" in the listed works, as we only recently have named our ontology "SOMA".}.
To conclude this section, we will elaborate on best practices in ontology engineering that we have established for collaborative development of SOMA.

\begin{figure}[t]
\tikzset{
      mynode/.style={outer,fill=pastel1,draw=pastel3,text badly centered,text width=2.0cm,inner sep=3pt,minimum height=0.5cm},
      mylarge/.style={mynode,text width=5.0cm},
      myedge/.style={implies,pastel5,thick},
      mylabel/.style={black!80,font={\scriptsize\sffamily}},
      mytitle/.style={font=\scriptsize\sffamily\bf,draw=pastel5,fill=pastel1,inner sep=0.04cm},
}
\begin{multicols}{2}
\begin{tikzpicture}
  \node[mynode,physical] (EV) {objects};
  \node[mynode,social,right=0.8cm of EV] (PL) {plans};
  \node[inner sep=0,yshift=1.8cm] at ($(EV)!0.5!(PL)$) (EXEC) {
    \includegraphics[width=5.3cm,trim=0mm 30mm 0mm 42mm, clip]{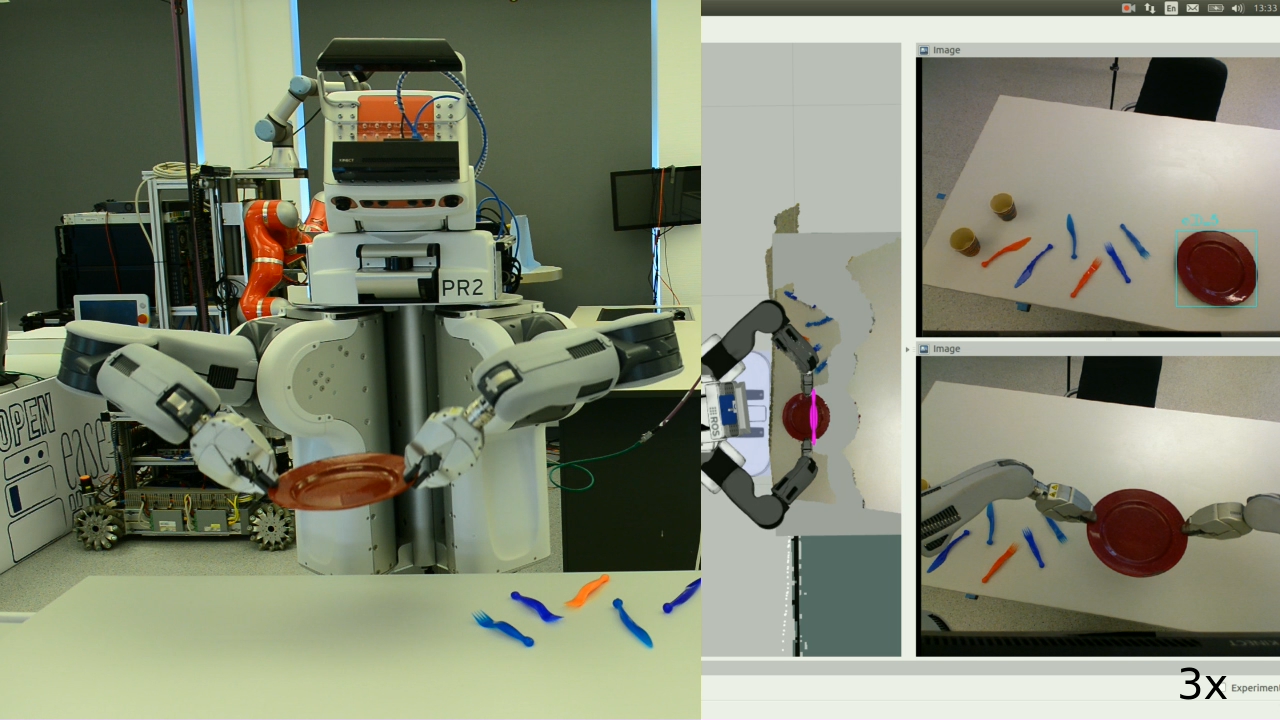}
  };
  \node[mytitle,physical] at (EXEC.north) {Execution};
  \node[mylarge,yshift=-1.0cm] at ($(EV)!0.5!(PL)$) (DIS) {dispositional matching};
  \path[] (EXEC.south -| EV.north) edge[myedge] node[left,mylabel,xshift=-0.1cm]{perception}
          (EV.north);
  \path[] (EXEC.south -| PL.north) edge[myedge] node[right,mylabel,xshift=0.1cm,yshift=0.02cm]{context}
          (PL.north);
  \path[] (EV.south) edge[myedge] node[left,mylabel,xshift=-0.1cm]{facts}
          (EV.south |- DIS.north);
  \path[] (PL.south) edge[myedge] node[right,mylabel,xshift=0.1cm]{facts}
          (PL.south |- DIS.north);
  \path[] (DIS) edge[myedge] node[right,mylabel,yshift=-0.5cm]{selection}
          (EXEC);
\end{tikzpicture}
\par 
\begin{tikzpicture}
  \node[mynode,physical] (EV) {events};
  \node[mynode,social,right=0.8cm of EV] (PL) {plans};
  \node[inner sep=0,yshift=1.8cm] at ($(EV)!0.5!(PL)$) (VR) {
    \includegraphics[width=5.3cm,trim=0mm 0mm 0mm 6mm, clip]{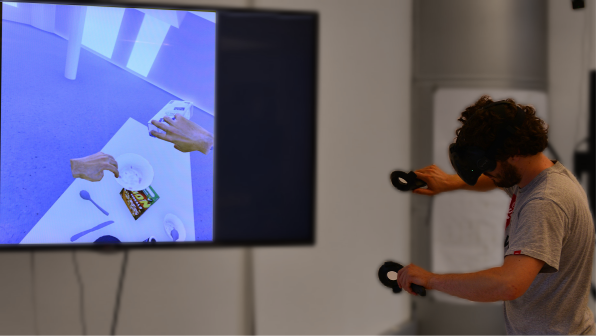}
  };
  \node[mytitle,physical] at (VR.north) {Observation};
  \node[mylarge,yshift=-1.0cm] at ($(EV)!0.5!(PL)$) (MAP) {activity parsing};
  \path[] (VR.south -| EV.north) edge[myedge] node[left,mylabel,xshift=-0.1cm]{tokens}
          (EV.north);
  \path[] (EV.south) edge[myedge] node[left,mylabel,xshift=-0.1cm]{facts}
          (EV.south |- MAP.north);
  \path[] (PL.south) edge[myedge] node[right,mylabel,xshift=0.1cm]{facts}
          (PL.south |- MAP.north);
  \draw[pastel5,thick,dashed,latex-latex] (EV) -- (PL);
\end{tikzpicture}
\end{multicols}
\vspace{-0.6cm}
\caption{SOMA is used for object selection (left); and for activity interpretation (right).}
\label{fig:application}
\end{figure}
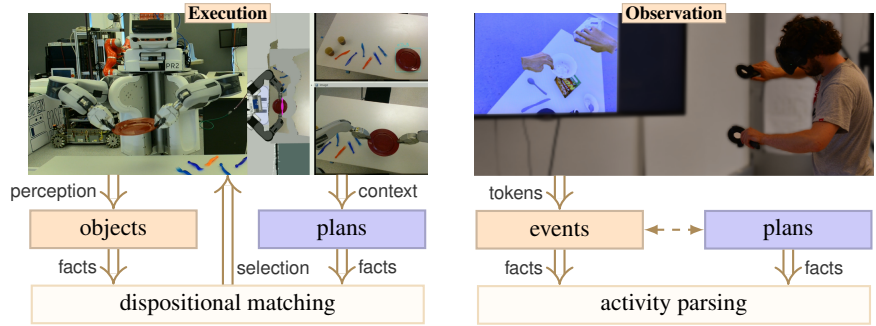

Each manipulation task involves a set of objects that are classified by certain roles during the task. The roles of a task are conceptualized by constraints on the class of objects that may play the role during task execution.
We have shown examples on how dispositional qualities in SOMA can be axiomatized that allow to define general plans that refer to dispositional classes of objects~\cite{bessler2020ecai}.
This dispositional object selection is depicted in the left part of Figure~\ref{fig:application}.

Grounding task parameters often requires predictive models which can be trained over instances of successful performance.
Such experiential knowledge is in particular useful to learn context-dependent \emph{plan specializations}. That is, how the parameters of the plan can be constrained within the scope of some context to reduce the search space of parameter selection during plan execution. The learning problem is then defined with respect to a contextual pattern, and experiential samples are only considered when their contextualization matches this pattern.
We have demonstrated this capability in another work where a robot learns to execute a general fetch and place plan based on experience acquired through the execution of more constrained tasks~\cite{koralewski19specialization}.

Learning mechanisms often require large amounts of training data. This raises the question how huge amounts of experiential knowledge can be acquired. One modality for acquisition is observation of other agents where the intention of the agent is unknown.
In order to acquire experiential knowledge from observations, we have developed an activity parser that is used to find possible interpretations for observed patterns of occurrences such as that objects get into contact with each other, or that the state of an object changed~\cite{bessler19vr}.
The grammar used by the parser is a library of plans represented using SOMA.
This is depicted in the right part of Figure~\ref{fig:application}.
Each plan describes the structure of a task by axiomatizing occurrences of steps using Allen's interval algebra such that one formal plan can match many sequences of observed events.

In another work, we have provided more details about how the social context in SOMA can be grounded in data structures of a game engine~\cite{haidu18gameengine}.
The game engine implements an immersive virtual reality environment with photo realistic rendering and state of the art physics engine. Users perform object manipulation tasks while interactions, states, and motions are monitored, and used as tokens by the activity parser.
The goal is to generate large amounts of experiential knowledge that implicitly encodes common sense and naive physics knowledge, and to learn from these episodes.

Our modeling of tasks
also helps to disambiguate vague natural language commands a human user might give to a robot. SOMA defines concepts such as \texttt{Binding} and \texttt{ConditionalSuccedence} which allows us to model how tasks relate to and depend on one another, and thus define execution contexts containing not just information about a task's parametrization, but also information about what other tasks it should enable. We use such execution contexts to set up simulation scenarios in which to test task executions and thus select among several interpretations of a vague natural language command~\cite{bateman2019}.

Currently, SOMA contains more than 3000 asserted facts across all previously mentioned modules.
To make downstream work easier, we combine all modules into one overarching ontology. However, an ontology of SOMA's size is difficult to maintain: It is always possible that changes render the ontology inconsistent.
To prevent these kinds of errors, we employ strict workflows and adhere to best practices in software engineering. We use the version control system git to make it easy to share small patches of the ontology among our team and apply the patches after at least one member of our team reviewed the changes. Even the best review might not catch subtle inconsistencies. To catch such errors, we automatically deploy the ontology with each change set to a continuous integration server that tests for inconsistencies by using an OWL reasoner.

\section{Conclusion}


In this paper, we have proposed SOMA, a novel activity ontology for robotic agents that combines several established ontology design patterns with models of human cognition to cope with the challenges of everyday activities in a general manner. The SOMA ontology has been validated through a set of use-cases that demonstrate its applicability for reasoning about tasks in the scope of autonomous robot control. This is done by representing both the physical and the social context of an activity, and by establishing relationships between both contextualizations. These representations are used by robotic agents to fill knowledge gaps in general plans applicable to many situations, and to generate context-specific behavior. Another aspect is that agents may as well use SOMA for the representation of observed actions performed by another agent, and for reasoning about their social contextualization.
We believe that such an expressive activity representation is an important vehicle for transforming robots from just performing a task to mastering the corresponding activity.


\bibliographystyle{splncs}
\bibliography{ecaw}

\begin{thebibliography}{10}

\bibitem{lenat91}
Lenat, D.B., Feigenbaum, E.A.:
\newblock On the thresholds of knowledge.
\newblock Artificial Intelligence \textbf{47}(1) (1991)  185 -- 250

\bibitem{anderson95}
Anderson, J.E.:
\newblock Constraint-Directed Improvisation for Everyday Activities.
\newblock PhD thesis, The University of Manitoba (Canada) (1995) AAINN99082.

\bibitem{Olivares19}
Olivares-Alarcos, A., Beßler, D., Khamis, A., Gonçalves, P., Habib, M.,
  Bermejo, J., Barreto, M., Diab, M., Rosell, J., Quintas, J., Olszewska, J.,
  Nakawala, H., Pignaton~de Freitas, E., Gyrard, A., Borgo, S., Alenyà, G.,
  Beetz, M., Li, H.:
\newblock A review and comparison of ontology-based approaches to robot
  autonomy.
\newblock The Knowledge Engineering Review \textbf{34} (12 2019)

\bibitem{Patel2012}
Patel, R., Hedelind, M., Lozan{-}Villegas, P.:
\newblock Enabling robots in small-part assembly lines: The "rosetta approach"
  - an industrial perspective.
\newblock In: {ROBOTIK}, VDE-Verlag (2012)

\bibitem{Malec2013}
Malec, J., Nilsson, K., Bruyninckx, H.:
\newblock Describing assembly tasks in declarative way.
\newblock In: {IEEE/ICRA} Workshop on Semantics. (2013)

\bibitem{Stenmark2015}
Stenmark, M., Malec, J., Nilsson, K., Robertsson, A.:
\newblock On distributed knowledge bases for robotized small-batch assembly.
\newblock IEEE Transactions on Automation Science and Engineering
  \textbf{12}(2) (2015)  519--528

\bibitem{Balakirsky2013}
Balakirsky, S., Kootbally, Z., Kramer, T., Pietromartire, A., Schlenoff, C.,
  Gupta, S.:
\newblock Knowledge driven robotics for kitting applications.
\newblock Robot. Auton. Syst. \textbf{61}(11) (November 2013)  1205--1214

\bibitem{Balakirsky2015}
Balakirsky, S.:
\newblock Ontology based action planning and verification for agile
  manufacturing.
\newblock Robotics and Computer-Integrated Manufacturing \textbf{33}(Supplement
  C) (2015)  21 -- 28 Special Issue on Knowledge Driven Robotics and
  Manufacturing.

\bibitem{PolydorosGRNK16}
Polydoros, A.S., Gro{\ss}mann, B., Rovida, F., Nalpantidis, L., Kr{\"{u}}ger,
  V.:
\newblock Accurate and versatile automation of industrial kitting operations
  with skiros.
\newblock In: Towards Autonomous Robotic Systems - 17th Annual Conference
  {(TAROS)}. (2016)  255--268

\bibitem{Kootbally2015}
Kootbally, Z., Schlenoff, C., Lawler, C., Kramer, T., Gupta, S.:
\newblock Towards robust assembly with knowledge representation for the
  planning domain definition language (pddl).
\newblock Robot. Comput.-Integr. Manuf. \textbf{33}(C) (June 2015)  42--55

\bibitem{PerzyloSPKRK16}
Perzylo, A., Somani, N., Profanter, S., Kessler, I., Rickert, M., Knoll, A.:
\newblock Intuitive instruction of industrial robots: Semantic process
  descriptions for small lot production.
\newblock In: {IEEE/RSJ} International Conference on Intelligent Robots and
  Systems {(IROS)}. (2016)  2293--2300

\bibitem{tenorth2015representations}
Tenorth, M., Beetz, M.:
\newblock Representations for robot knowledge in the knowrob framework.
\newblock Artificial Intelligence (2015)

\bibitem{schlenoff2012ieee}
Schlenoff, C., Prestes, E., Madhavan, R., Goncalves, P., Li, H., Balakirsky,
  S., Kramer, T., Miguelanez, E.:
\newblock An {IEEE} standard ontology for robotics and automation.
\newblock In: {IEEE} Int. Conf. on Intelligent Robots and Systems (IROS).
  (2012)  1337--1342

\bibitem{PRESTES20131193}
Prestes, E., Carbonera, J.L., Fiorini, S.R., Jorge, V.A.M., Abel, M., Madhavan,
  R., Locoro, A., Goncalves, P., Barreto, M.E., Habib, M., Chibani, A.,
  Gérard, S., Amirat, Y., Schlenoff, C.:
\newblock Towards a core ontology for robotics and automation.
\newblock Robotics and Autonomous Systems \textbf{61}(11) (2013)  1193 -- 1204
  Ubiquitous Robotics.

\bibitem{Fiorini2015}
Fiorini, S.R., Carbonera, J.L., Gon\c{c}alves, P., Jorge, V.A., Rey, V.F.,
  Haidegger, T., Abel, M., Redfield, S.A., Balakirsky, S., Ragavan, V., Li, H.,
  Schlenoff, C., Prestes, E.:
\newblock Extensions to the core ontology for robotics and automation.
\newblock Robot. Comput.-Integr. Manuf. \textbf{33}(C) (June 2015)  3--11

\bibitem{DOLCE2003}
Masolo, C., Borgo, S., Gangemi, A., Guarino, N., Oltramari, A.:
\newblock {WonderWeb} deliverable {D18} ontology library (final).
\newblock Technical report, IST Project 2001-33052 WonderWeb: Ontology
  Infrastructure for the Semantic Web (2003)

\bibitem{Baker98}
Baker, C.F., Fillmore, C.J., Lowe, J.B.:
\newblock The berkeley framenet project.
\newblock In: Proceedings of the 36th Annual Meeting of the Association for
  Computational Linguistics and 17th International Conference on Computational
  Linguistics - Volume 1. ACL ’98/COLING ’98, USA, Association for
  Computational Linguistics (1998)  86–90

\bibitem{gibson1979ecological}
Gibson, J.J.:
\newblock The Ecological Approach to Visual Perception.
\newblock Psychology Press Classic Editions (1979)

\bibitem{Wan18}
Yamanobe, N., Wan, W., Ramirez-Alpizar, I.G., Petit, D., Tsuji, T., Akizuki,
  S., Hashimoto, M., Nagata, K., Harada, K.:
\newblock A brief review of affordance in robotic manipulation research.
\newblock Journal of the Robotics Society of Japan \textbf{36} (01 2018)
  327--337

\bibitem{Ortmann10}
Ortmann, J., Kuhn, W.:
\newblock Affordances as qualities.
\newblock In: Proceedings of the 2010 Conference on Formal Ontology in
  Information Systems: Proceedings of the Sixth International Conference (FOIS
  2010), Amsterdam, The Netherlands, The Netherlands, IOS Press (2010)
  117--130

\bibitem{moralez2016affordance}
Moralez, L.A.:
\newblock Affordance ontology: towards a unified description of affordances as
  events.
\newblock Res. Cogitans \textbf{7}(1) (2016)  35--45

\bibitem{turvey1992ecological}
Turvey, M.T.:
\newblock Ecological foundations of cognition: Invariants of perception and
  action.
\newblock American Psychological Association (1992)

\bibitem{bessler2020ecai}
Be{\ss}ler, D., Porzel, R., Pomarlan, M., Beetz, M., Malaka, R., Bateman, J.:
\newblock {A Formal Model of Affordances for Flexible Robotic Task Execution}.
\newblock In: European Conference on Artificial Intelligence (ECAI). (2020)
  accepted for publication.

\bibitem{flanagan06control}
Flanagan, J.R., Bowman, M.C., Johansson, R.S.:
\newblock Control strategies in object manipulation tasks.
\newblock Current Opinion in Neurobiology (2006)

\bibitem{Allen83}
Allen, J.F.:
\newblock Maintaining knowledge about temporal intervals.
\newblock In: In Communications of the ACM, Volume 26 Issue 11. (1983)

\bibitem{Talmy2000}
Talmy, L.:
\newblock Toward a Cognitive Semantics. Volume 2: Typology and Process in
  Concept Structuring.
\newblock Language, Speech, and Communication. MIT Press, Cambridge, MA (2000)

\bibitem{koralewski19specialization}
Koralewski, S., Kazhoyan, G., Beetz, M.:
\newblock Self-specialization of general robot plans based on experience.
\newblock Robotics and Automation Letters (2019)

\bibitem{bessler19vr}
Be{\ss}ler, D., Porzel, R., Mihai, P., Beetz, M.:
\newblock Foundational models for manipulation activity parsing.
\newblock In Jung, T., tom Dieck, M.C., Rauschnabel, P.A., eds.: Augmented
  Reality and Virtual Reality.
\newblock Springer (2019) 978-3-030-37869-1.

\bibitem{haidu18gameengine}
Haidu, A., Be{\ss}ler, D., Bozcuoglu, A.K., Beetz, M.:
\newblock Knowrob-sim — game engine-enabled knowledge processing for
  cognition-enabled robot control.
\newblock In: International Conference on Intelligent Robots and Systems
  (IROS), Madrid, Spain, IEEE (2018)

\bibitem{bateman2019}
Bateman, J., Pomarlan, M., Kazhoyan, G.:
\newblock Embodied contextualization: Towards a multistratal ontological
  treatment.
\newblock Applied Ontology \textbf{14}(4) (2019)  379--413

\end{thebibliography}

\end{document}